\documentclass[acmsmall,screen,nonacm]{acmart}

\IfFileExists{upquote.sty}{\usepackage{upquote}}{}
\IfFileExists{microtype.sty}{
  \usepackage[]{microtype}
  \UseMicrotypeSet[protrusion]{basicmath} 
}{}
\makeatletter
\@ifundefined{KOMAClassName}{
  \IfFileExists{parskip.sty}{%
    \usepackage{parskip}
  }{
    \setlength{\parindent}{0pt}
    \setlength{\parskip}{6pt plus 2pt minus 1pt}}
}{
  \KOMAoptions{parskip=half}}
\makeatother

\PassOptionsToPackage{unicode}{hyperref}
\PassOptionsToPackage{hyphens}{url}
\PassOptionsToPackage{dvipsnames,svgnames,x11names}{xcolor}

\IfFileExists{bookmark.sty}{\usepackage{bookmark}}{\usepackage{hyperref}}

\usepackage{longtable,booktabs,array}
\usepackage{calc} 
\usepackage{etoolbox}
\makeatletter
\patchcmd\longtable{\par}{\if@noskipsec\mbox{}\fi\par}{}{}
\makeatother
\IfFileExists{footnotehyper.sty}{\usepackage{footnotehyper}}{\usepackage{footnote}}
\makesavenoteenv{longtable}
\usepackage{graphicx}
\makeatletter
\newsavebox\pandoc@box
\newcommand*\pandocbounded[1]{
  \sbox\pandoc@box{#1}%
  \Gscale@div\@tempa{\textheight}{\dimexpr\ht\pandoc@box+\dp\pandoc@box\relax}%
  \Gscale@div\@tempb{\linewidth}{\wd\pandoc@box}%
  \ifdim\@tempb\p@<\@tempa\p@\let\@tempa\@tempb\fi
  \ifdim\@tempa\p@<\p@\scalebox{\@tempa}{\usebox\pandoc@box}%
  \else\usebox{\pandoc@box}%
  \fi%
}
\def\fps@figure{htbp}
\makeatother

\providecommand{\tightlist}{%
  \setlength{\itemsep}{0pt}\setlength{\parskip}{0pt}}

\usepackage[]{natbib}
\usepackage{multirow}
\usepackage{array}
\usepackage{placeins}
\usepackage{tikz}
\usetikzlibrary{positioning,arrows.meta}
\newcommand{\mfull}{\tikz[baseline=-0.6ex]\fill (0,0) circle (0.75ex);}
\newcommand{\mhalf}{\tikz[baseline=-0.6ex]{\draw (0,0) circle (0.75ex); \fill (0,0) -- (90:0.75ex) arc (90:270:0.75ex) -- cycle;}}
\newcommand{\mnone}{\tikz[baseline=-0.6ex]\draw (0,0) circle (0.75ex);}
\newcommand{\mnt}{\tikz[baseline=-0.6ex]{\draw[dashed, line width=0.5pt] (0,0) circle (0.75ex); \node[font=\tiny] at (0,0) {?};}}

\definecolor{mypink}{RGB}{219, 48, 122}
\makeatletter
\@ifpackageloaded{caption}{}{\usepackage{caption}}
\AtBeginDocument{%
\ifdefined\contentsname
  \renewcommand*\contentsname{Table of contents}
\else
  \newcommand\contentsname{Table of contents}
\fi
\ifdefined\listfigurename
  \renewcommand*\listfigurename{List of Figures}
\else
  \newcommand\listfigurename{List of Figures}
\fi
\ifdefined\listtablename
  \renewcommand*\listtablename{List of Tables}
\else
  \newcommand\listtablename{List of Tables}
\fi
\ifdefined\figurename
  \renewcommand*\figurename{Figure}
\else
  \newcommand\figurename{Figure}
\fi
\ifdefined\tablename
  \renewcommand*\tablename{Table}
\else
  \newcommand\tablename{Table}
\fi
}
\@ifpackageloaded{float}{}{\usepackage{float}}
\floatstyle{ruled}
\@ifundefined{c@chapter}{\newfloat{codelisting}{h}{lop}}{\newfloat{codelisting}{h}{lop}[chapter]}
\floatname{codelisting}{Listing}

\makeatother
\makeatletter
\@ifpackageloaded{caption}{}{\usepackage{caption}}
\@ifpackageloaded{subcaption}{}{\usepackage{subcaption}}
\makeatother

\hypersetup{
  pdftitle={SoK: Grading Open-Text Legal Answers{,} from Law School Exams to Trustworthy Rubric Judges},
  pdfauthor={Sebastian Nagl},
  colorlinks=true,
  linkcolor={blue},
  filecolor={Maroon},
  citecolor={Blue},
  urlcolor={red},
  pdfcreator={LaTeX via pandoc, via quarto}}

\AtBeginDocument{%
  }

\setcopyright{none}
\copyrightyear{2027}
\acmYear{2027}
\acmDOI{XXXXXXX.XXXXXXX}

\acmConference[CSLAW '27]{the 6th ACM Symposium on Computer Science and
Law}{March 23--25, 2027}{Chicago, IL, USA}
\acmISBN{978-x-xxxx-xxxx-x/YYYY/MM}
\begin{document}

\title[SoK: Grading Open-Text Legal Answers]{SoK: Grading Open-Text
Legal Answers, from Law School Exams to Trustworthy Rubric Judges}
  \author{Sebastian Nagl}
  \orcid{0009-0009-6321-9775}
  \email{sebastian.nagl@tum.de}
            \affiliation{%
                  \institution{Technical University of Munich (TUM)}
                          \city{Munich}
                          \country{Germany}
              }

\begin{abstract}
Law students are examined through open-text answers: case solutions,
essays and memos that apply the law to a set of facts. Such answers are
expensive to grade, human graders disagree widely on them, and large
language models (LLMs) now offer to grade them at scale, both in
teaching and in the benchmarks that test legal AI. This paper
systematizes how open-text legal answers can be evaluated. We derive
eight desiderata from educational measurement and from the law that
governs exams, among them reasons, credit for defensible alternative
solutions, robustness to gaming and human oversight. We then compare
eight evaluation paradigms against them, from unstructured human grading
to LLM judges with instance-specific rubrics and pairwise comparison.
Three findings stand out. A rubric written for the task is the strongest
lever for agreement between an LLM judge and human graders, but studies
measure that agreement against different references and rarely report
how far the human graders agree with each other. The central validation
gap lies where examination law is most demanding: defensible alternative
solutions, follow-on errors, and the examiner's correction of model
judgments. Rubric-centred grading is the most auditable family, because
it makes the standard, its application and the final decision checkable.
From this we derive the trustworthy rubric judge: an instance rubric
that credits arguments and alternative paths, evidence checked in code,
probes, repeated runs, calibration against independent human grades,
logging, and an examiner who reads every script. Today the evidence
supports such a judge for formative feedback and for benchmarks, and as
an assistant to the examiner, not a replacement, where credit is at
stake.    
\end{abstract}



\maketitle

\setlength{\parskip}{-0.1pt}

\section{Introduction}\label{sec-intro}

Legal education tests a skill that is hard to measure: applying the law
to facts in writing. A German student receives a fact pattern
(\emph{Sachverhalt}) and writes an opinion in appraisal style
(\emph{Gutachtenstil}); a US student analyses a hypothetical in IRAC
form (issue, rule, application, conclusion); a bar candidate drafts a
memo or a judgment. The exam case stands in for the file a lawyer or a
court receives, and the opinion rehearses the analysis that precedes
advice or a decision. These answers run to many pages, admit more than
one defensible solution, and are graded by experts who often disagree.
The same German exam can receive grades several points apart on the
18-point scale \citep{hufeld2024}, and when 17 US law professors graded
the same essays without a scoring guide, a single professor's grade
reached an intraclass correlation (ICC) of .58 with the others
\citep{klein1968nature}. Grading is also slow, so students get little
feedback before the graded exam, although individual feedback measurably
improves their grades \citep{schwarcz2017impact}.

Large language models (LLMs) have changed both sides of this problem.
They write answers that pass law exams
\citep{choi2023chatgpt, katz2024gpt4, martinez2025reevaluating}, so
legal AI needs benchmarks that grade open text. And they offer to grade
answers themselves, as an \emph{LLM judge}
\citep{zheng2023mtbench, gu2024survey}. Since 2025, legal benchmarks and
teaching tools have converged on one design: an expert writes a rubric
for each task, and an LLM checks an answer against it
\citep{shi2026plawbench, chen2026lexrubric, drozdov2026legalresearchbench, zubaer2026gradelegal, cope2026grading}.
At the same time, LLM judges are known to drift between runs
\citep{haldar2025rating}, to favour long answers
\citep{saito2023verbosity}, to obey instructions embedded in the answer
they grade \citep{li2026important}, and to reward whatever an optimizer
learns about their disposition \citep{elganayni2026exploiting}.
Examination law, data protection law and the EU AI Act set further
limits on what a machine may decide about a candidate
\citep{bverwg2019_6C19_18, cjeu2023schufa, aiact2024}.

The literature on these questions is scattered across law reviews,
educational measurement, natural language processing (NLP) and court
decisions. This paper brings them together. In NLP, an automatic grader
is a predictor of the human grade, validated by its agreement with that
grade. In examination law, a grade is the application of a stated
standard to an answer that may legitimately differ from the reference,
and it must be justified. We adopt the legal definition: a grading
method must apply a stated standard and justify its result, not only
agree with human graders. Our question is \emph{which way of evaluating
open-text legal answers the evidence supports, for which use, and what
must still be shown before it can be trusted}. The evidence is of three
kinds: direct studies of automatic graders of legal answers, rubric
grading and rater research from other domains, and the law that governs
grading (Section~\ref{sec-method}). Answering the question means asking
what a grade should measure (Section~\ref{sec-desiderata}), how that
performance is translated into criteria and points
(Section~\ref{sec-design}), and who answers for that translation and can
correct it in the single case (Section~\ref{sec-trust}).

Our contributions are:

\begin{enumerate}
\def\labelenumi{\arabic{enumi}.}
\tightlist
\item
  \textbf{Desiderata.} Eight requirements for evaluating open-text legal
  answers, from validity theory
  \citep{messick1995validity, kane2013validating} and from examination,
  data protection and AI law (Section~\ref{sec-desiderata}, Table
  \ref{tab:requirements}).
\item
  \textbf{A taxonomy of eight paradigms}, from unstructured human
  grading to LLM judges with instance-specific rubrics and pairwise
  comparison, with the evidence for and against each
  (Section~\ref{sec-paradigms}, Table \ref{tab:paradigms}).
\item
  \textbf{A systematization of legal evaluations} of automatic graders:
  task-specific content is the strongest lever, agreement is measured
  against different references, and validation is often thin
  (Section~\ref{sec-legal}, Table \ref{tab:studies}).
\item
  \textbf{A design space of legal rubrics}, from the unit of credit to
  the key that maps points to grades, with the evidence and law behind
  each choice (Section~\ref{sec-design}).
\item
  \textbf{The trustworthy rubric judge}: seven safeguards, each tied to
  a weakness, a legal requirement and the test that remains, with the
  uses the evidence supports today (Section~\ref{sec-trust},
  Section~\ref{sec-uses}, Section~\ref{sec-open}).
\end{enumerate}

An analytic rubric written for the task is the strongest lever for
agreement between an automatic grader and human graders. No architecture
wins on every axis: when attorneys ranked answers written at three
quality levels, comparing them in pairs recovered the order better than
rubric scores did \citep{yang2026judgmentbench}; multiple-choice formats
are more reliable; and no rubric judge is yet robust to manipulation. We
argue that rubric-centred grading, grading that checks each answer
against criteria stated in advance, should still be the default for
auditable legal assessment. Grading is auditable when the standard, its
application to each script and the final decision can each be checked; a
rubric externalizes the standard, makes each criterion application
contestable and leaves the decision with the examiner. These are design
advantages motivated by examination law, not a claim that the law
mandates one architecture. We propose the \emph{trustworthy rubric
judge}, a rubric judge wrapped in seven safeguards, and state the tests
it must pass before it grades exams for credit. A companion study, in
preparation, builds this judge for German law exams
\citep{nagl2027transformation} (Section~\ref{sec-open}).

\section{Background: open-text legal answers and how humans grade
them}\label{sec-background}

\emph{The tasks.} We call an answer \emph{open-text} when the candidate
writes free prose that applies law to facts and no answer key can list
every acceptable text. In Germany the core format is the case solution
(\emph{Falllösung}): an opinion that states each question, defines the
legal requirements, subsumes the facts and reaches a result, graded on
the statutory scale from 0 to 18 points \citep{jurprnotskv1981}.
Practice tasks include memos, contract drafting and research answers
\citep{schwarcz2026aipowered, drozdov2026legalresearchbench} and draft
judgments \citep{karp2025llm}. Our core case is the exam; the other
tasks differ in the performance measured, in who sets the standard and
in what is at stake, and we do not claim that the findings transfer.

\emph{Human graders disagree.} In the 1968 study of 17 US law
professors, the mean of all their grades reached a reliability of .96
against .58 for one professor, so single graders and panel means are
different yardsticks \citep{klein1968nature}. On 15 scripts of a German
beginner-level mock exam, each graded by 15 or 16 correctors who had the
model solution but no point scheme, the highest and lowest grade for a
script lay 6.47 points apart on average \citep{hufeld2024}, and in
BenGER, a benchmark of German case solutions curated on an open platform
\citep{nagl2026benger}, three blind reviewers reached a single-rater
ICC(2,1) of .63 \citep{nagl2026bengerdata}. Grades in the German state
exams also differ by gender and by the origin of a candidate's name
\citep{towfigh2014benotung, towfigh2018geschlechts}, and by the
\emph{Land} in which the exam is taken \citep{kahler2017hangen}. Grades
rise with length \citep{augustineadams2016speed}, legible handwriting
\citep{greifeneder2010writing} and legal jargon \citep{linn1972nature},
and graders are drawn towards a grade they have already seen
\citep{vidalrodeiro2007agreement, attali2011sequential}.

\emph{Structure helps, but not alone.} An analytic rubric breaks the
expected answer into criteria, each with points or levels; it is
\emph{instance-specific} when it is written for one task rather than for
a type of task \citep{dawson2017assessment}. Rubrics raise agreement
between human raters, especially when they are analytic and
task-specific and come with exemplars or rater training
\citep{jonsson2007use}, and rubrics used formatively improve student
performance \citep{panadero2023effects}. In German law, some correctors
note what each part of the expected solution is worth
\citep{frenzel2011korrektur}, and a few make this binding in a grading
sheet (\emph{Bewertungsbogen}), described for take-home papers
\citep{heidebach2015pruefen}, which narrows the spread between
correctors (Table \ref{tab:studies}). Yet rubrics do not remove rater
effects: raters still differ in severity
\citep{woitschach2018efectos, engelhard1994examining}, read the same
criteria differently \citep{bloxham2016lets}, and without moderation
shared criteria did not make law marking consistent
\citep{hanlon2004exploring}. High-stakes testing surrounds the rubric
with machinery: calibration on common scripts, hidden scripts with known
scores to catch drifting severity
\citep{olson2019best, congdon2000stability}, independent double grading
and adjudication \citep{light2024grading}, which the US NextGen bar exam
now adopts for its written answers
\citep{light2024grading, bok2024grading, ncbe2026nextgen}.

\section{Method of this systematization}\label{sec-method}

\emph{Scope.} We draw on three bodies of sources and keep them apart in
the text, since they carry different weight. Direct evidence is work
that grades or evaluates open-text answers to legal tasks, by humans or
machines. Transferred evidence is work from other domains that tests
rubric-based LLM grading against human graders, and the measurement
literature on rubrics and raters. Normative sources are the legal rules
that govern grading: German examination law as the worked example, the
EU General Data Protection Regulation (GDPR) and AI Act, and US cases on
academic judgment. We exclude closed formats except where compared with
open text, and supervised essay-scoring models, for which we found no
use on law exams.

\emph{Search.} We searched the ACL Anthology, arXiv, Google Scholar,
SSRN, the ACM Digital Library, legal education journals and German and
EU case-law databases up to 12 October 2026, in English and German, with
keywords on law exam grading, legal benchmarks, rubric and checklist
evaluation, LLM-as-a-judge, automated essay and short-answer scoring,
and exam law, and followed references in both directions; language
models assisted by proposing and retrieving candidate sources. A legal
evaluation enters Table \ref{tab:studies} if it grades open-text legal
answers automatically and reports how, or whether, the grader was
compared with humans; reuse of another study's scripts or rubrics is
marked, and so are preprints.

\emph{Prior surveys.} Surveys and critiques cover LLM judges
\citep{gu2024survey, chehbouni2025neither}, essay and short-answer
scoring outside law
\citep{ke2019automated, burrows2015eras, gao2024automatic, li2025agreement}
and legal benchmarks by task and metric \citep{hu2026evaluation}; none
relates grading methods to the law that governs grading or treats the
rubric as the object of design.

\emph{Coding.} For each legal evaluation we recorded the paradigm
(Section~\ref{sec-paradigms}), the source of the rubric, what the grader
was compared with, and the reported agreement of the grader and of the
human graders (Table \ref{tab:studies}). Agreement is reported as the
source reports it: Pearson's \(r\) or Spearman's \(\rho\), Cohen's
\(\kappa\) or its quadratic weighted form (QWK), and the intraclass
correlation (ICC) for rater reliability. We rated each paradigm against
each desideratum as supported, partly or under conditions, contradicted,
or not tested, from the evidence discussed in
Section~\ref{sec-paradigms}. The basis of a mark is a comparison on
legal answers, evidence from another domain, a legal source, or our
reading.

\section{Desiderata}\label{sec-desiderata}

What makes an evaluation good depends on what it is used for. We follow
argument-based validation: a score is valid for a use when each
inference from the answer to the decision is backed by evidence
\citep{kane2013validating, aera2014standards}. The law adds requirements
of its own, which we state first.

\subsection{What the law asks of grading}\label{sec-law}

Table \ref{tab:requirements} lists the requirements.

\begin{table}[t]
\caption{What the law asks of grading open-text exam answers (L1 to L4 German examination law, L5 to L7 EU law), the implementation we propose (Section~\ref{sec-trust}) and what is still open.}
\label{tab:requirements}
\centering
\scriptsize\setlength{\tabcolsep}{3pt}
\begin{tabular*}{\linewidth}{@{\extracolsep{\fill}}l>{\raggedright\arraybackslash}p{0.27\linewidth}>{\raggedright\arraybackslash}p{0.17\linewidth}>{\raggedright\arraybackslash}p{0.26\linewidth}>{\raggedright\arraybackslash}p{0.19\linewidth}@{}}
\toprule
 & Requirement & Source & Proposed implementation & Open \\
\midrule
L1 & Equal standards: each examiner's standards for a task apply to every script of that task alike & \cite[p.~52]{bverfge84_34}; \cite[para.~8]{bverwg2018_6B71_17} & one versioned rubric per examiner and task; corrections applied to every script; pinned judge version & untested across scripts and versions \\
L2 & Answer latitude: a defensible, consistently argued solution must not be marked wrong; an error is not counted twice & \cite[p.~79]{bverfge84_59}; \cite[p.~55]{bverfge84_34}; \cite{vgbraunschweig2018}; \cite{vglueneburg2024} & credit for arguments; separate paths; later steps judged against the candidate's own decision; unforeseen paths flagged & no grader probed with alternatives or follow-on errors \\
L3 & Reasons: the essential reasons for the grade, not the reason for every point & \cite{bfh2014viib158}; \cite[para.~23]{bverwg2019_6C19_18}; Art.~15(1)(h) GDPR \cite{cjeu2025dunbradstreet} & a verdict and a quoted passage per criterion, beyond the legal minimum & a quote shows the passage, not that the criterion is met \\
L4 & Personal evaluation and reconsideration by the examiner & \cite{bverwg2002_6C7_02}; \cite{bverwg2019_6C19_18} & the examiner reads every script and decides; the judge proposes and flags & correction of confident wrong proposals untested \\
L5 & No decision based solely on automated processing with legal or similarly significant effect, unless necessary for a contract, authorized by law with safeguards, or based on explicit consent & Art.~22 GDPR \cite{gdpr2016}; \cite{cjeu2023schufa} & substantive review, not adoption of a score as given & what counts as substantive review is unsettled \\
L6 & AI that evaluates learning outcomes is high-risk; providers owe accuracy, robustness and logging, deployers competent oversight, logs and candidate information, public bodies also a fundamental-rights assessment (duties from 2 Dec 2027) & \cite{aiact2024}, Annex III, point 3(b), Arts.~9 to 15, 26, 27, 86; \cite{omnibus2026} & documented validation, logging, examiner oversight with authority to override, probes against manipulation & scope of Arts.~2(6) and 6(3) for prototypes \\
L7 & Exam scripts and examiner comments are personal data & \cite{cjeu2017nowak} & legal basis, processing contract or local models, pseudonymized inputs, access and deletion rules & pseudonymization alone is insufficient \\
\bottomrule
\end{tabular*}
\end{table}

\emph{German examination law.} Exams that control access to a profession
fall under the freedom of occupation (Art. 12(1) of the Basic Law, GG).
Examiners have a margin of appreciation. Each examiner sets their own
standards for a task and must apply them to every script of that task
alike \citetext{\citealp[p.~52]{bverfge84_34}; \citealp[para.
8]{bverwg2018_6B71_17}}; the law does not demand one scheme shared by
all examiners. On judicial review of an exam decision, courts check in
full whether a defensible answer was marked wrong; examination-specific
judgments of persuasiveness, structure and consistency are reviewed only
for the limits of that discretion, a complete and correct reading of the
script, equal application of the examiner's standards, and freedom from
improper considerations and contradictions \citep[paras. 9 to
11]{bverwg2018_6B71_17}. The candidate has an \emph{answer latitude}
(\emph{Antwortspielraum}): a solution that is defensible
(\emph{vertretbar}) and argued consistently with weighty reasons must
not be marked as wrong
\citetext{\citealp[p.~79]{bverfge84_59}; \citealp[p.~55]{bverfge84_34}}.
Latitude protects the result, not the mark: a defensible but weakly
argued solution may earn fewer points than a stronger one. An error must
also not be counted twice. A follow-on error (\emph{Folgefehler}) is
reasoning that is wrong or open to criticism only because the candidate
took an earlier decision on the path of the solution, and penalizing it
again is an impermissible double counting (\emph{Doppelverwertung})
\citep{vgbraunschweig2018}; how much a follow-on error still weighs is
left to the examiner's judgment \citep{vglueneburg2024}. The candidate
is owed the essential reasons for the grade \citep[para.
23]{bverwg2019_6C19_18}, but has no right to margin notes that justify
every single point, and a point scheme is an aid the examiner may depart
from, not a binding rule \citep{bfh2014viib158}. On substantiated
objections the examiners themselves must reconsider their evaluation; no
other body may do it for them, and written reasons do not replace this
reconsideration \citep{bverwg2019_6C19_18}. The examiner must personally
and directly acquaint themselves with each script and evaluate it
independently; an open second correction, in which the second examiner
sees the first grade, is not forbidden by federal constitutional law
\citep{bverwg2002_6C7_02}. US courts likewise defer to the academic
judgment of faculty \citep{horowitz1978, ewing1985}, so a professor's
evaluation must stay substantive when a model assists.

\emph{Data protection.} A candidate's answers and the examiner's
comments on them are personal data \citep{cjeu2017nowak}, so sending
scripts to a model provider needs a legal basis and a processing
contract. Art. 22 GDPR gives a person the right not to be subject to a
decision based solely on automated processing that has legal or
similarly significant effects, unless the decision is necessary for a
contract, authorized by Union or Member State law with suitable
safeguards, or based on explicit consent \citep{gdpr2016}. The Court of
Justice of the European Union (CJEU) has held that a score is such a
decision when a third party's decision draws strongly on it
\citep{cjeu2023schufa}, so an examiner who adopts a model grade as given
risks a solely automated decision; human involvement must be meaningful
\citep{colonna2024teachers}, and the person may require an explanation
of the procedure and principles applied \citep{cjeu2025dunbradstreet}.

\emph{AI regulation.} The AI Act classifies AI systems intended to
evaluate learning outcomes as high-risk (Annex III, point 3(b)), which
brings the provider duties of risk management, data governance,
technical documentation, logging, transparency, human oversight,
accuracy and robustness against manipulation (Arts. 9 to 15)
\citep{aiact2024}. A university that uses such a system is a deployer:
it must assign oversight to persons with the competence and authority to
intervene, keep the logs and inform candidates that the system is used
on them (Art. 26); a public university must also assess the impact on
fundamental rights before first use (Art. 27); and a candidate may
demand an explanation of the system's role in the decision (Art. 86).
Systems developed and put into service solely for scientific research
fall outside the Act (Art. 2(6)), and a system that only prepares a
human evaluation may escape the high-risk class (Art. 6(3)), but a grade
proposal that the examiner reviews shapes the decision and is unlikely
to escape it. The Digital Omnibus postponed the provider and deployer
duties for Annex III systems to 2 December 2027 \citep{omnibus2026} (law
as of October 2026).

\subsection{Eight desiderata}\label{sec-desiderata-list}

From the measurement literature and the requirements above we derive
eight desiderata.

\begin{itemize}
\tightlist
\item
  \textbf{D1 Construct validity.} The score supports the interpretation
  it is used for: it reflects the legal skill the task tests, such as
  spotting issues, stating rules, subsuming facts and weighing
  arguments, and not task-irrelevant features such as length, surface
  style or the grader's identity
  \citep{messick1995validity, williamson2012framework}. Fairness is part
  of this: the score should not move with task-irrelevant features of a
  candidate's language, such as phrasing or dialect.
\item
  \textbf{D2 Reliability.} The score changes little with the grader, the
  run or the occasion. In generalizability theory these are separate
  sources of error \citep{brennan2001generalizability}.
\item
  \textbf{D3 Reasons.} The grade comes with reasons that let the
  candidate understand and contest it (L3, L5); a reason per criterion
  and passage is the form we propose (Section~\ref{sec-trust}).
\item
  \textbf{D4 Defensible alternatives.} A defensible solution that
  departs from the reference is not marked wrong, and an error is not
  counted again in the steps that depend on it (L2).
\item
  \textbf{D5 Robustness.} The score cannot be raised by padding, copying
  the case, citing fake authority or embedded instructions
  \citep{powers2002stumping, chen2024humans, li2026important}.
\item
  \textbf{D6 Cost and scale.} The method can grade many answers fast
  enough for feedback and benchmarks.
\item
  \textbf{D7 Formative value.} The output tells a student what was
  missing and why \citep{hattie2007power}.
\item
  \textbf{D8 Examiner control.} The examiner's standards apply to every
  script alike (L1); the examiner, not the system, decides and
  reconsiders (L4, L5); and the process is documented as the AI Act's
  oversight duties require (L6).
\end{itemize}

Whether a deployment is lawful depends on context and procedure, not on
the method alone. The desiderata overlap: D3, D4 and D8, where the law
adds most, all favour designs that make the standard explicit, so a
paradigm's count of marks is not a score. Agreement with human graders,
the usual yardstick in NLP, is evidence for D1 and D2, not proof: humans
disagree, an automatic grader can match a noisy reference by chance
\citep{elangovan2025beyond}, and consistency is not validity
\citep{han2026agreement}.

\section{Eight evaluation paradigms}\label{sec-paradigms}

We distinguish eight paradigms by who grades (human or model) and what
the grade is anchored to (nothing, a reference answer, general criteria,
criteria written for the task, or another answer). They are the
configurations found in the literature, not disjoint classes: the design
space behind them has more axes, and real systems mix them, as a
pairwise judge with a rubric does \citep{cope2026grading}. Table
\ref{tab:paradigms} rates each paradigm on the desiderata.

\begin{table}[t]
\caption{Evaluation paradigms against the desiderata of Section~\ref{sec-desiderata}.}
\label{tab:paradigms}
\centering
\footnotesize
\begin{tabular*}{\linewidth}{@{\extracolsep{\fill}}llcccccccc@{}}
\toprule
 & Paradigm & \rotatebox{70}{Valid} & \rotatebox{70}{Reliable} & \rotatebox{70}{Reasons} & \rotatebox{70}{Alternatives} & \rotatebox{70}{Robust} & \rotatebox{70}{Scale} & \rotatebox{70}{Feedback} & \rotatebox{70}{Oversight} \\
\midrule
P1 & Human grading, unstructured & \mhalf & \mnone & \mhalf & \mfull & \mhalf & \mnone & \mhalf & \mfull \\
P2 & Human grading, analytic rubric & \mfull & \mhalf & \mfull & \mhalf & \mhalf & \mnone & \mfull & \mfull \\
P3 & Closed formats & \mnone & \mfull & \mnone & \mnone & \mfull & \mfull & \mnone & \mhalf \\
P4 & Similarity to a reference & \mnone & \mfull & \mnone & \mnone & \mnone & \mfull & \mnone & \mnone \\
P5 & LLM judge, reference answer & \mhalf & \mhalf & \mhalf & \mnone & \mnone & \mfull & \mhalf & \mhalf \\
P6 & LLM judge, generic criteria & \mnone & \mhalf & \mhalf & \mnt & \mnone & \mfull & \mhalf & \mhalf \\
P7 & LLM judge, instance rubric & \mfull & \mhalf & \mfull & \mnt & \mnone & \mfull & \mhalf & \mhalf \\
P8 & Pairwise comparison & \mhalf & \mhalf & \mnone & \mnt & \mhalf & \mhalf & \mnone & \mnone \\
\bottomrule
\end{tabular*}
\par\smallskip
\parbox{\linewidth}{\footnotesize \mfull{} supported; \mhalf{} partly or under conditions; \mnone{} contradicted; \mnt{} not tested. Marks rate the evidence, not what a design could achieve; Section~\ref{sec-paradigms} discusses the evidence behind the marks.}
\end{table}

\emph{P1 Unstructured human grading (``vibe grading'').} An expert reads
the answer and gives a grade, perhaps with margin comments. It is the
legal default and the reference most studies validate against, and an
expert can recognize a defensible alternative that no key foresaw. But
the two studies that measured it put single-rater reliability at
intraclass correlations of .58 and .63
\citep{klein1968nature, nagl2026bengerdata}, correctors' grades spread
widely \citep{hufeld2024}, and length, handwriting, identity and prior
grades leak into the grade (Section~\ref{sec-background}).

\emph{P2 Human grading with an analytic rubric.} A rubric makes the
standard explicit, cuts the spread between human graders
\citep{strecker2026ki, jonsson2007use} and gives students criteria to
learn from \citep{panadero2023effects, sparrow2004describing}. Its
weaknesses are cost and the remaining rater effects, and it credits
alternatives only if the rubric lists them or the grader departs from
it.

\emph{P3 Closed formats.} Multiple choice, classification and extraction
tasks are graded exactly and cheaply, which is why most legal NLP
benchmarks use them
\citep{chalkidis2022lexglue, guha2023legalbench, zhong2020jecqa, fei2024lawbench}.
But they measure a different construct. With GPT-4's help, law students
improved by 29 percentile points on multiple-choice questions and not at
all on essays \citep{choi2025ai}, and GPT-4's bar exam percentile is
lowest on the essays, about the 42nd against the 62nd overall
\citep{martinez2025reevaluating}.

\emph{P4 Similarity to a reference.} Metrics such as ROUGE, BLEU and
BERTScore compare the answer with a reference text
\citep{lin2004rouge, papineni2002bleu, zhang2020bertscore}. On legal
summaries, experts' scores correlated with ROUGE at .13 to .21 and with
BERTScore at .07 \citep{shukla2022legal}; the model with the highest
ROUGE on case-law analyses produced the highest rate of unsupported
citations \citep{hou2025clerc}, and near-identical ROUGE on German court
press releases hid very different expert ranks
\citep{nagl2025courtpressger}. Similarity cannot give reasons, punishes
a defensible alternative path for using other words, and rewards copying
the reference.

\emph{P5 LLM judge with a reference answer.} The judge compares the
answer with a reference answer written by an expert, as in LEXam
\citep{fan2026lexam}. Without a reference, judges agree with experts
only on questions they can answer themselves \citep{krumdick2025no}, and
judging accuracy tracks solving accuracy \citep{tan2025judgebench}. With
a reference, LEXam's judge passed the alternative annotator test against
three experts who agreed with each other at a quadratic weighted kappa
of .49 \citep{calderon2025alternative, fan2026lexam}. A single reference
also fixes one path. In BenGER, the judge disagreed with human graders
most on presentation and on results reached by other reasoning paths
than the reference solution's \citep{nagl2026bengerdata}, which is
exactly the answer latitude the law protects (L2). Such verdicts are
also judge-relative: in our earlier work, prompts optimized against a
strict judge on LEXam overfit to that judge's disposition, while prompts
optimized against a lenient judge transferred better
\citep{elganayni2026exploiting}.

\emph{P6 LLM judge with generic criteria.} The judge scores every answer
on the same generic dimensions \citep{liu2023geval}. In law this fails
without task-specific guidance. Open models grading German criminal-law
case solutions without a sample solution or rubric reached correlations
near zero \citep{wendlinger2024suitability}, and a G-Eval-style
correctness judge, LeMAJ's baseline (Table \ref{tab:studies}),
correlated with legal experts at .02 to .08 \citep{enguehard2025lemaj}.
Generic dimensions combined with a reference fare better: BenGER's
judge, which scores each case solution on ten general criteria of German
correction practice against the reference solution, agreed with the mean
of three blind human graders at \(r\) = .76, within the graders' own
disagreement \citep{nagl2026bengerdata}, and a judge with general
criteria ranked systems for German court press releases much as experts
did \citep{nagl2025courtpressger}. What makes them work is the
task-specific content (P7).

\emph{P7 LLM judge with an instance-specific rubric.} An expert writes,
for each task, the criteria an answer must meet, often as atomic
yes-or-no items, and the judge checks each criterion. Outside law, this
design gives judges high agreement with experts per criterion. On
HealthBench's 34 consensus criteria, GPT-4.1's agreement with physicians
fell within the range of the physicians' agreement with each other
\citep{arora2025healthbench}, and similarly on expert rubrics in other
fields \citep{starace2025paperbench, wang2026profbench}; controlled
comparisons favour instance-specific criteria over general ones, over a
plain prompt and over free grading
\citep{kim2025biggen, wei2025rocketeval, du2025deepresearch, liu2026aiassisted, lee2024applying}.
In law, the within-domain comparisons point the same way, with caveats
(Table \ref{tab:studies}). On German exams, giving GPT-5 the grading
scheme raised its agreement with the human graders more than giving it
the sample solution did, and both together raised it most
\citep{zubaer2026gradelegal}. The gain is not universal: the same
additions lowered the agreement of GPT-5.1, and the scripts are 71
take-home solutions and 16 timed mock-exam scripts reused from earlier
studies \citep{wendlinger2024suitability, strecker2026ki}. On four US
law-school finals, GPT-5 with the professor's rubric tracked the
professor's grades closely and better than with a bare prompt; a score
per rubric element and pairwise comparison with the rubric added nothing
\citep{cope2026grading}. On the one question a professor regraded, the
two human gradings were closer to each other than to the model, a gap
the authors do not find significant. On attorney-written binary rubrics
for legal research, judges from three providers agreed with the
attorneys' majority label somewhat better than the attorneys agreed with
each other, a comparison that is not like for like
\citep{drozdov2026legalresearchbench}.

The weaknesses are real. Expert rubrics cost time: building each
PLawBench item, rubric included, took about three expert hours
\citep{shi2026plawbench}. Generated rubrics fall short of expert ones
\citep{wang2026generating, dhole2026rubricrag, zhou2026rubricbench}, and
LLM-written elaborations of human rubrics lowered agreement
\citep{sharma2025researchrubrics}. Finer label schemes lower agreement
\citep{deng2025rubric, sharma2025researchrubrics}
(Section~\ref{sec-design}). Judges also side with the majority. In a
Thai bar-exam study of reference-based judges, where the grading
regulation was silent, 22 of 26 LLM judges scored in or near the
majority reading of three examiners and none reproduced the defensible
minority reading; the dissent was among graders about the same answer,
not an alternative path taken by the candidate
\citep{akarajaradwong2026thai}. And a rubric does not catch every
fundamental error. In a Polish study of the exam for the
public-procurement appeals board, a judge given the committee's criteria
and the correct legal analysis gave three model-written judgments 85 to
90 of 100 points; the committee had failed all three for cardinal
errors, with 8 to 37 points \citep{karp2025llm}. Grading against a
rubric also invites manipulation: instructions embedded in an answer
succeeded in 56.9\% of attacks on average across rubric-based graders
\citep{li2026important}.

\emph{P8 Pairwise comparison.} The grader compares two answers and picks
the better one; many comparisons yield a ranking, as in Chatbot Arena
\citep{zheng2023mtbench} or comparative judgment. This paradigm is the
strongest challenge to rubrics, for one question: ordering quality. In
JudgmentBench, practising attorneys graded answers to 30 BigLaw Bench
tasks written at three quality levels. Their pairwise judgments
recovered the order of the levels almost perfectly, their rubric scores
barely, and the rubric raters agreed with each other only moderately; a
GPT-5.4 autograder tracked the attorneys' orderings better with pairwise
judgments than with the rubric \citep{yang2026judgmentbench} (Table
\ref{tab:studies}). Pairwise ranking also beat absolute scoring for
essays \citep{shibata2025lces} and resisted universal adversarial
phrases better \citep{raina2024llm}. JudgmentBench's levels were induced
by prompting and moved all dimensions together, and its rubrics were not
validated against graded answers; real answers differ in which issues
they treat, which is what rubrics capture. Pairwise judges show position
and length bias \citep{shi2024judging, dubois2024length}. Above all, a
bare preference is relative: it ranks a script among others but gives no
criterion-referenced reasons, no feedback, and no grade that holds when
the cohort changes. That conflicts with the duty to give reasons (L3)
and with criterion-referenced grading, towards which even the bar exam
is moving \citep{light2024grading}. The objection is to the protocol,
not to comparison as such: a pairwise judge with the rubric matched
absolute rubric grading on US exams \citep{cope2026grading}, and
anchored comparisons with a reason per criterion are untested. As used
today, P8 suits leaderboards better than exams.

\section{Systematization of legal evaluations}\label{sec-legal}

Table \ref{tab:studies} lists the automatic graders of open-text legal
answers we found, what each was compared with, and whether it graded
student scripts or model answers. Three patterns stand out.

\begin{table*}[t]
\caption{Automatic graders of open-text legal answers, by paradigm. \emph{Judge}: agreement with the reference in \emph{Compared with}, model stability marked as such; \emph{Human}: reliability of the human grades, between or within graders; \emph{reused}: data or instruments shared with another cited study; n/r: not reported. QWK: quadratic weighted kappa; SPA: soft pairwise accuracy; IAA, IRR: inter-annotator agreement and reliability; MAE: mean absolute error; $W$ and $\tau$: Kendall's; $\alpha$: Krippendorff's.}
\label{tab:studies}
\centering
\scriptsize\setlength{\tabcolsep}{2pt}\renewcommand{\arraystretch}{0.95}
\begin{tabular*}{\linewidth}{@{\extracolsep{\fill}}>{\raggedright\arraybackslash}p{0.25\linewidth}l>{\raggedright\arraybackslash}p{0.16\linewidth}>{\raggedright\arraybackslash}p{0.12\linewidth}>{\raggedright\arraybackslash}p{0.19\linewidth}>{\raggedright\arraybackslash}p{0.17\linewidth}@{}}
\toprule
Study and setting & P & Rubric & Compared with & Judge & Human \\
\midrule
Wendlinger et al. \cite{wendlinger2024suitability}: German criminal-law take-home, student scripts & P6 & none (few-shot examples) & research assistants' grades & near zero or negative & n/r \\
LeMAJ \cite{enguehard2025lemaj}: legal QA, model answers with constructed errors & P6 & none; answer split into legal data points & legal experts & correctness $r$ .26 (own data), .70 (LegalBench) & IAA correctness .77; .88 with data points \\
Karp et al. \cite{karp2025llm}: Polish procurement appeals-board exam, 3 model judgments & P5/P7 & committee's criteria and solution key & official committee & 85 to 90 of 100 for failed judgments (committee 8 to 37) & n/r \\
BenGER \cite{nagl2026bengerdata}: German case solutions, 15 tasks, human-written & P5/P6 & reference solution, 10 general criteria & mean of 3 blind reviewers & $r$ .76; pass/fail $\kappa$ .60 & ICC(2,1) .63, single rater (45 answers, 15 by models) \\
CourtPressGER \cite{nagl2025courtpressger}: German court press releases, model outputs & P6 & 5 general criteria & expert ranking of 10 systems & $\rho$ .94 (systems) & n/r \\
SLDS \cite{rolshoven2025unlocking}: Swiss headnotes, model outputs & P5 & reference headnote, generic categories & expert ratings & $\rho$ .26 (ROUGE-L .26) & n/r \\
LEXam \cite{fan2026lexam}: Swiss law exams, model answers & P5 & professor's reference answer & 3 experts, alternative annotator test & passes & QWK .49 \\
GreekBarBench \cite{chlapanis2025greekbarbench}: Greek bar exam, model answers & P5/P7 & expert spans on the reference & aggregated expert scores, system pairs & soft pairwise accuracy .86 (system level) & SPA .85 (annotator vs. annotator); $\alpha$ .74 \\
OAB-Bench \cite{pires2025automatic}: Brazilian bar exam, 3 passing candidate scripts & P7 & official point tables & official scores & item MAE .04 to .28 & n/r \\
Magis-Bench \cite{pires2026magisbench}: Brazilian magistrate exams, model answers & P7 & official examiner rubrics & other LLM judges only & $W$ .98 & n/r \\
GradeLegal \cite{zubaer2026gradelegal}: German criminal and public law, student scripts; reused & P7 & sample solution, grading scheme & instructor (public), assistants (criminal) & QWK .60 (crim.), .91 (public); .31, .58 with neither & QWK .91 (public; instructor vs. assistants) \\
Strecker et al. \cite{strecker2026ki}: German public-law mock exam, student scripts & P2/P7 & examiner's raw-point scheme & instructor's grade; weighted human mean & deviation mostly lower with scheme; grades compressed & spread 5.25 vs. 2.94 points with scheme (two groups) \\
Cope et al. \cite{cope2026grading}: US finals, 4 exams, 205 student scripts & P7/P8 & professor's rubric & professor's grade & rubric $r$ .78 to .93; open .66 to .80; per element, pairwise no better & one regraded question: humans closer to each other than to the model, not significant \\
Thai bar study \cite{akarajaradwong2026thai}: Thai bar essays, 15 human and model answers & P5/P6 & generic 8-tier regulation, gold answer & 3 examiners' single scores & 3 anchor judges $r$ .52 to .76; LLM-panel stability $\alpha$ .77 & panel $\alpha$ .36 (.75 where explicit) \\
KCL-Essay \cite{oh2026korean}: Korean bar-exam questions, model answers & P7 & LLM-generated from expert solution, 27\% attorney-reviewed & attorney scores & $r$ .8 (.7 on hard items) & $r$ .93 (easy items) \\
PLawBench \cite{shi2026plawbench}: Chinese practice tasks, model answers & P7 & about 12,500 expert items & at least 3 experts, sampled items & $r$ .72 to .86 by task type & $r$ .84 to .88 (expert vs. mean) \\
LexRubric \cite{chen2026lexrubric}: Chinese consultation and exams, model answers & P7 & 12,337 expert atomic criteria & experts, 50 instances; model ranking & item $\kappa$ .59; $\tau$ .80 & n/r \\
Legal Research Bench \cite{drozdov2026legalresearchbench}: US legal research, 413 questions, model answers & P7 & attorney-written binary items, reviewed & majority of 3 attorneys, 341 items & item $\kappa$ .71 to .73 & $\kappa$ .64 (attorney vs. attorney) \\
JudgmentBench \cite{yang2026judgmentbench}: BigLaw Bench, model outputs at 3 constructed levels; rubrics reused & P7/P8 & vendor rubrics by attorneys & attorneys' orderings & GPT-5.4 $\rho$ .20 (rubric), .38 (pairwise) & attorneys $\rho$ .91 pairwise, .15 rubric; IRR .44 \\
\bottomrule
\end{tabular*}
\end{table*}

\emph{The field has moved from P3 and P4 to P7.} Legal benchmarks long
reduced law to closed formats or scored generation with overlap or
keyword metrics
\citep{chalkidis2022lexglue, fei2024lawbench, li2024lexeval, li2025legalagentbench}.
Since 2025, the leading benchmarks and the practitioners' own
evaluations have adopted instance-specific rubrics, written or seeded by
experts, with an LLM judge
\citep{shi2026plawbench, chen2026lexrubric, oh2026korean, drozdov2026legalresearchbench, grupen2026lab, vals2025vlair}.

\emph{Task-specific content is the strongest lever, within limits.}
Within the same domain, a scheme or rubric raised the agreement of the
same model, with the caveats of Section~\ref{sec-paradigms}.
Decomposition can also start from the answer instead of the expected
performance: LeMAJ splits a model's answer into legal data points and
checks each against the source material without a reference answer
\citep{enguehard2025lemaj}. The two are not the same. A rubric fixes in
advance what counts as performance and who is accountable for that
choice. Decomposing the answer only organizes the checking, omissions
included; it supplies neither weights nor a grade.

\emph{Validation is often thin, and human-level claims depend on the
unit of comparison.} Several rubric benchmarks report agreement between
judges but not with humans \citep{pires2026magisbench}, validate on
three candidate scripts that all passed \citep{pires2025automatic}, or
publish no agreement at all
\citep{harvey2024biglawbench, grupen2026lab}. Agreement between judges
is not validity \citep{rao2026jev, han2026agreement}. Few report how far
their human graders agree with each other. Where both figures are
reported, the comparison is rarely like for like: the judge is scored
against a majority or a mean, the humans against each other
\citep{drozdov2026legalresearchbench, chlapanis2025greekbarbench, nagl2026bengerdata},
and the alternative annotator test is the exception (S5). A panel of
three judges agrees with itself better than three examiners agree with
each other, which does not show that it agrees with them
\citep{akarajaradwong2026thai}. The sharpest contrasts come from quality
levels constructed by prompting \citep{yang2026judgmentbench}. And
correlation is not accuracy: a correlation of one is compatible with
every grade three points too high, harmless for a ranking and decisive
for pass or fail. Table \ref{tab:studies} shows judges that agree with
human graders, not grades that hold up in the single case. The next two
sections turn to a design that could.

\section{The design space of legal rubrics}\label{sec-design}

The evidence of Section~\ref{sec-paradigms} and Section~\ref{sec-legal}
favours instance-specific rubrics, but ``rubric'' covers very different
instruments \citep{dawson2017assessment}. Table \ref{tab:design}
summarizes the choices that any grading sheet for open-text legal
answers makes, whether a human or a model applies it.

\begin{table}[t]
\caption{The design space of legal grading rubrics; the option we propose is listed first.}
\label{tab:design}
\centering
\scriptsize\setlength{\tabcolsep}{3pt}
\begin{tabular*}{\linewidth}{@{\extracolsep{\fill}}>{\raggedright\arraybackslash}p{0.18\linewidth}>{\raggedright\arraybackslash}p{0.46\linewidth}>{\raggedright\arraybackslash}p{0.32\linewidth}@{}}
\toprule
Dimension & Options & Evidence and law \\
\midrule
Unit of credit & argument or requirement met; issue spotted; bare result & answer latitude (L2); experts agree almost perfectly on binary criteria \\
Scale of a criterion & binary; few levels; free points per step & binary criteria agree best with experts; coarser labels also discard information \\
Structure & all points on steps (absolute or as shares of the total); steps plus a bounded, reasoned component; steps plus a criteria-bound pool; steps plus a free pool & a free pool cannot be checked; a criteria-bound pool needs stated criteria and a fixed share \\
Method and form & within each section; separate block & within sections stays tied to content; a block needs stated criteria and a fixed share \\
Alternative solutions & outcome-neutral criteria and paths that record why a step is absent; flag unforeseen paths; single reference & L2; equal performance, not equal length; the examiner's rule for the total \\
Follow-on errors & standard per criterion (correct law or own earlier decision); none & not counted twice; the weight is the examiner's (L2) \\
Cardinal errors & defined in advance (minus points or a lower grade); left to the grader & examiners disagree; a deduction is itself a grading decision \\
Points to grade & examiner's key applied by code; key generated or applied by a model & the key to the legal scale is the examiner's (L1, L4); models mis-add \\
Author and timing & examiner, before grading; model draft with marked provenance, reviewed by the examiner; model alone & equal standards, versioned (L1); generated rubrics fall short \\
\bottomrule
\end{tabular*}
\end{table}

\emph{What earns credit.} Graders of law essays reward the treatment of
major issues more than minor ones \citep{linn1972nature}, and a correct
result without the authority behind it is a different achievement from a
correct, authority-grounded answer \citep{liao2026knowing}. A rubric
that credits bare results penalizes defensible alternatives and rewards
guessing; one that credits the reasoning (the requirement stated, the
facts subsumed and the argument weighed) fits both legal method and the
answer latitude (L2).

\emph{Granularity and scale.} Yes-or-no verdicts agree better than
partial-credit or finer label schemes
\citep{sharma2025researchrubrics, deng2025rubric}, and on binary expert
criteria, experts agree almost perfectly \citep{wang2026profbench}. But
splitting criteria into ever smaller items does not always help
\citep{kortemeyer2024grading, liu2026aiassisted}. Agreement is not the
construct: coarser labels raise agreement partly by discarding
information. The granularity should follow what the step tests, and the
states \emph{not met}, \emph{not applicable}, \emph{uncertain} and
\emph{check failed} must stay apart even where the criterion is binary.
The middle ground is a step for each expected legal performance, with a
few binary requirements per step; on US exams the most granular rubric,
not the most granular output, gave the highest agreement, a hypothesis
from four exams \citep{cope2026grading}.

\emph{Structure and weights.} German point schemes come in three types.
In the first, every point belongs to a step of the expected solution,
either as absolute points per step
\citep{heidebach2015pruefen, frenzel2011korrektur} or as a percentage of
the total for each item of the outline, so that no conversion key is
needed \citep[pp.~12 f.]{strecker2026ki}. In the second, a fixed share
goes to steps and a pool is bound to named criteria, typically method
points (\emph{Strukturpunkte}) for structure, appraisal style and
language, for which 15 to 25\% is proposed
\citep[p.~14]{strecker2026ki}; the same criteria can instead be credited
within each section. In the third, the pool is free: the grader
distributes it on an overall impression. A free pool is tied to nothing
a second grader or a model could check and reintroduces the variance the
sheet was meant to remove, so it has no place in a rubric judge. A
bounded overall component that must name what it rewards, for instance
up to a tenth of the points for the weighting of the issues or for an
unforeseen but sound solution, awarded only with a written reason, can
be stated, limited and reviewed; it belongs in the design space as a
documented hybrid (Section~\ref{sec-open}).

\emph{Alternatives and follow-on errors.} Where a dispute changes only a
result or its reasons, a criterion can credit the argument whatever the
result. Where another defensible decision removes or adds steps, the
rubric needs a separate path whose steps are weighted for equal
performance, not for equal length, and answers that take a path no one
foresaw should go to the examiner. A worked example shows the three
cases a rubric must keep apart. Suppose question A decides whether step
B is reached. If the candidate defensibly answers A in the negative,
step B does not arise and neither earns nor loses points, unless the
task asks for an auxiliary opinion (\emph{Hilfsgutachten}), in which
case B is still expected. If the candidate denies A wrongly, what
follows from that decision is a follow-on error, judged against the
candidate's own decision with the weight of the wrong decision left to
the examiner, while a new error inside B counts on its own. The rubric
must record why a step is absent, or it credits work not shown, or
penalizes a lawful path for omitting steps it did not need; and a judge
must not pick, from a contradictory answer, the path that scores best.
Marking B not applicable does not settle the total: if A carries four
points and B six, a denominator of ten penalizes the lawful path, and
rescaling to four turns a shorter performance into a full score.
Alternative paths need an examiner's rule for comparable scoring
opportunities and for the total.

\emph{Cardinal errors.} Examiners disagree on which errors alone should
take a script below the pass mark, and the literature asks each sheet to
define them in advance, with minus points or a grade below the computed
one \citep[pp.~15 f.]{strecker2026ki}. The Polish study of
Section~\ref{sec-paradigms} shows the stakes: a judge with the
committee's criteria and solution key gave near-full marks to judgments
the committee had failed for cardinal errors \citep{karp2025llm}.
Whether such a rule would have caught them was not tested; a deduction
is itself a grading decision that needs a reason.

\emph{From points to grade.} The key that maps points to the legal grade
scale is part of the examiner's evaluation: the performance required for
a passing grade cannot be fixed rigidly without regard to average
results \citep[pp.~51 f.]{bverfge84_34}, and once set, it applies to
every script alike \citep[para. 8]{bverwg2018_6B71_17}. Since models
mis-add \citep{kortemeyer2024grading, baeumel2025lookahead}, the sum and
the key belong in code, not in the judge's output.

\emph{Who writes the rubric, and when.} The examiner sets the standard
(L4). The rubric is fixed before grading; a defect found while grading
is corrected in a new version, the change is documented, and every
script graded under the old version is regraded on the changed criteria
(L1). Who drafts the rubric is a separate question. A model can draft
it, but drafting is more than extracting the reference solution: the
solution does not say which steps earn which points, what can be
expected under exam conditions, which alternatives are equal or which
errors weigh most. A draft should mark, per criterion, whether it
restates the task, the solution or the examiner's notes, adds an
alternative, proposes a weight or assumes something the examiner must
confirm, and the examiner must review and adopt it
(Section~\ref{sec-open}).

\emph{The rubric we propose.} Table \ref{tab:design} lists our choice
first in each row: credit for arguments and requirements met, not for
bare results (L2); binary criteria with the four states above kept
apart; every point on a step of the expected solution, method and form
credited inside each section so that each point stays tied to checkable
content; paths of their own for alternatives, follow-on errors judged
against the candidate's own earlier decision, and cardinal errors
defined in advance; the key from points to grade set by the examiner and
applied by code; the rubric written or reviewed by the examiner before
grading and versioned. The next section wraps the judge that applies it
in the safeguards the remaining gaps call for.

\section{The trustworthy rubric judge}\label{sec-trust}

\begin{figure}[t]
\centering
\includegraphics[width=\linewidth]{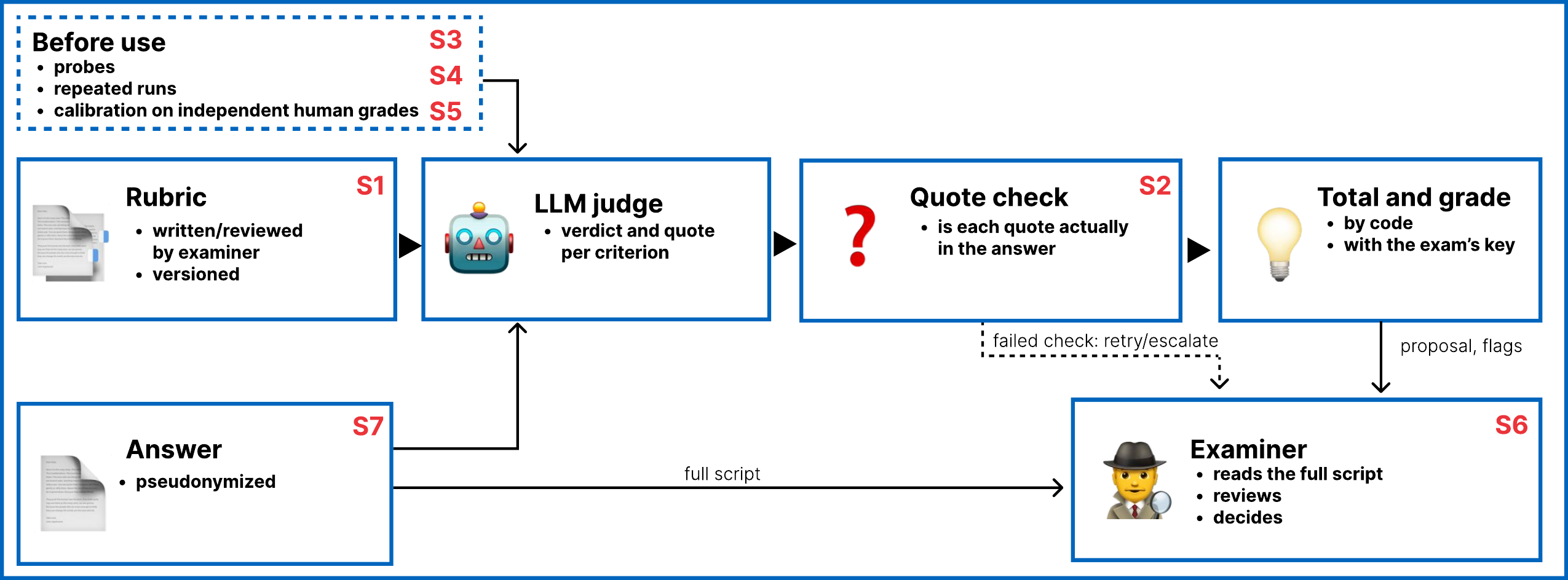}
\Description{Flow diagram. A versioned rubric written or reviewed by the examiner and a pseudonymized answer go to an LLM judge that returns a verdict and a quote per criterion. Code checks each quote against the answer, then computes the total and grade. The result goes to the examiner as a proposal with flags; a failed check goes back to the judge or to the examiner. The examiner also receives the full script. Before use, the judge passes probes, repeated runs and calibration on human grades.}
\caption{The trustworthy rubric judge. The rubric and the examiner are human, the judge is a model, the quote check and the total are code, and the dashed box is a procedure of code and humans run before use. A failed quote check is a system error, never a deduction; every grade is logged (S7).}
\label{fig:judge}
\end{figure}

Table \ref{tab:paradigms} shows no paradigm without weaknesses. P7 leads
on validity, reasons and scale, but it is open to manipulation, its
reliability is not guaranteed, and whether it credits alternatives or
improves learning is untested. We call a rubric judge \emph{trustworthy}
when safeguards close these gaps far enough for the intended use, and we
propose seven (Figure \ref{fig:judge}): two fix what the judge works
with and produces (S1, S2), three test it before and during use (S3 to
S5), and two keep the examiner in charge (S6, S7). Each safeguard has
support in the literature, from law or from other domains; the
combination is new, and Section~\ref{sec-open} states the tests it must
pass. Most of them would also fit a reference-based or pairwise judge,
but only a rubric makes every reason checkable against a criterion. The
closest prior design outside law is RULERS, a rubric judge that locks
the rubric, checks quoted evidence in code and fits its scores to human
grades by regression \citep{hong2026rulers}. What is specific here is
the legal standard: alternative paths, follow-on errors, weights set by
the examiner and the examiner's decision.

\textbf{S1 An instance rubric of atomic criteria that credits
arguments.} The judge applies the rubric of Section~\ref{sec-design}.
The examiner writes or reviews the criteria for each task
\citep{arora2025healthbench, drozdov2026legalresearchbench}; credit goes
to arguments, alternatives have paths of their own, and follow-on errors
are judged on the candidate's own decisions (L2). The rubric is fixed
before grading and versioned, so that a correction reaches every script
(L1). The examiner sets the weights and code does the arithmetic
\citep{kortemeyer2024grading}. Where a model drafts the rubric, an
expert must review it, since generated rubrics fall short of expert
instance rubrics \citep{wang2026generating, dhole2026rubricrag}. No
automatic grader of legal answers has yet been probed with alternative
paths or follow-on errors; S3 makes both a standard probe.

\textbf{S2 Evidence for every credit, checked in code.} The judge must
quote the passage of the answer that earns each credit, and a program
verifies that the quote is in the answer. Three questions are distinct:
does the passage exist, does it mean in context what the judge says, and
does that performance justify the points? A string check answers only
the first. The second needs care: a candidate may state a rule only to
reject it, and an omission has no passage to quote, so it needs a reason
instead. An invalid quote is an error of the grading system, not of the
candidate: it triggers a repeated run, an abstention or the examiner,
never a deduction. The quote gives the examiner and the candidate a
checkable basis for each point (L3), beyond what the law requires
\citep[paras. 8 f.]{bfh2014viib158}; the second and third questions
remain the examiner's.

\textbf{S3 Probes before use.} Before grading real answers, judge and
rubric must pass answers with known outcomes. Fake answers come first:
an empty answer, an off-topic one and a copy of the facts must score
near zero, the reference solution must score high, and an answer with
embedded instructions must not gain. Controlled variants of one real
answer come second, each with its permitted change in score fixed in
advance: padding must not raise the score, a paraphrase that keeps the
legal performance must not change it, a defensible alternative path must
keep its credit, the removal of a decisive reason must lose it, an
appended instruction must change nothing, and an invented authority must
not gain. A follow-on error probe changes an upstream decision and keeps
later steps consistent with it: the score may change only as the
examiner's dependency rule says, and a new error inside a dependent step
must still be found. Known attacks give the adversarial probes
\citep{raina2024llm, zhao2025one, li2026important, milani2025when, chen2024humans, zheng2025cheating, powers2002stumping}.
Probes answer the AI Act's demands for testing before use and for
robustness (Arts. 9(8), 15(4) and (5), L6), but show only that known
attacks fail.

\textbf{S4 Repeated runs and variance decomposition.} A judge can grade
the same answer differently on another run
\citep{haldar2025rating, schroeder2024trust}. The AI Act asks for
declared accuracy metrics and consistent performance (Art. 15(1) and
(3), L6), and equal standards (L1) ask for the same. One agreement
coefficient cannot show either when the judge's error has several
sources. Generalizability theory separates these sources, such as runs,
judges and rubrics \citep{brennan2001generalizability}, and has been
applied to LLM raters \citep{song2026exploring}. The rubric matters most
where rubrics are generated: a poor rubric can force the same error on
every script, so that every grade is consistent and wrong; less spread
is not success. A minimal design grades a sample of scripts on several
runs, under each rubric version and by two judges. It reports the
variance from runs, rubrics and judges against the variance between
scripts, and counts wrong passes and wrong fails apart. The run and
rubric shares must be small before a grade is used. Self-consistency
does not prove validity
\citep{norman2026reliability, sunkavalli2026llm}, so S4 complements S5
and does not replace it.

\textbf{S5 Calibration against independent human grades.} The judge is
validated on a sample of answers graded blind and independently by
humans, and its agreement is read against the agreement among the
humans, for instance with the alternative annotator test
\citep{calderon2025alternative, nagl2026bengerdata}. The human grades
must be independent, since second graders who see the first grade agree
more \citep{vidalrodeiro2007agreement}. Where several grades are
defensible, validation should accept a set of grades rather than one:
judges selected against forced single labels performed up to 31\% worse
\citep{guerdan2025validating}. Calibration and the final test must use
different data, split by task rather than by script if the claim is
about new exams. Calibration must be repeated when the model changes:
model updates shifted judge severity by up to 133 of 1,000 points
\citep{sunkavalli2026llm}. Calibration can mean two things. Improving
how the judge applies the criteria, through prompts, examples or quote
checks, leaves the standard untouched. Fitting the criterion weights or
the point-to-grade key to human scores, as RULERS does by regression
\citep{hong2026rulers}, changes the standard itself; that is a grading
decision, and the examiner must review and adopt it (L4).

\textbf{S6 Escalation to the examiner.} The judge proposes, and the
examiner decides (L4, L5). For a graded exam the examiner must read
every script personally \citep{bverwg2002_6C7_02}, so the proposal is a
draft, not a filter. Confidence is not the only flag: answers on an
unforeseen path, contradictions within the answer, criteria the judge
could not apply, failed quote checks, zeroed credits and grades near the
pass mark are flagged whatever the judge's confidence, since a
confidently wrong grade is the failure that matters. Where the judge is
uncertain, it abstains and leaves the script to the examiner. Such
selective grading can hold the agreement with humans on the graded
scripts at a chosen level, for scripts of the kind it was calibrated on
\citep{jung2025trust}, and raised it on one physics exam
\citep{kortemeyer2025assessing}; the guarantee is an average over many
scripts and says nothing about one candidate. Second readers are drawn
towards a grade they have seen \citep{vidalrodeiro2007agreement}, so the
examiner should form a view before seeing the model's grade; this is a
method, not a legal duty (Section~\ref{sec-law}). Whether examiners
catch a confidently wrong proposal is the test this safeguard must pass
(Section~\ref{sec-open}).

\textbf{S7 Documentation and data protection.} Every grade records what
produced it: the rubric version, the judge and model version, the
prompts, the quotes and the checks. It also records the judge's original
proposal and every human change with its reason, classed as a corrected
criterion application, a rubric change or a reasoned individual
decision. This record serves the logging and transparency duties of the
AI Act (Arts. 12, 13 and 26(6), L6), and it is the basis for the
explanation of the procedure and principles actually applied that a
candidate may request where the grade is a solely automated decision
\citep[paras. 58 to 61]{cjeu2025dunbradstreet}. Pseudonymized scripts
and local models are building blocks, not a data protection concept: the
controller, the legal basis, access rights and retention, deletion, and
any external service each need a decision of their own (L7).

The safeguards have costs: in one calculus exam, about 70\% of items had
to go to humans to reach human accuracy
\citep{kortemeyer2025artificial}, so the judge assists human graders and
does not replace them.

\section{Which uses the evidence supports today}\label{sec-uses}

The uses differ in stakes and in who bears an error, and the evidence
supports them unequally.

\emph{Formative feedback.} Feedback on practice exams is where the case
is strongest. Rubric use and feedback improve learning
\citep{panadero2023effects, wisniewski2020power}, individual feedback
raised law students' grades \citep{schwarcz2017impact}, and the American
Bar Association (ABA) now requires a formative assessment in every
course of the first third of the J.D. \citep{aba2026standards}. Whether
model-written feedback improves learning in law is untested; students in
German law still tended to prefer tutor feedback
\citep{hackl2024kifeedback}, and generous model grades mislead, so
feedback should state its uncertainty.

\emph{Benchmarks of legal AI.} For comparing models, an expert-rubric
judge validated against legal experts is now the best-supported design
\citep{drozdov2026legalresearchbench, shi2026plawbench}.

\emph{A second reader in graded exams.} A rubric judge that flags
scripts where its total and the examiner's differ is plausible, and a
review of the largest discrepancies on US exams found errors by the
professor as well as by the model \citep{cope2026grading}, but it is
barely tested. German law requires the examiner to decide (L4). Since
the features of the procedure that shape the result, such as the number
of examiners and the handling of their differences, must be fixed by
rule \citep[paras. 15 to 17]{bverwg2019_6C19_18}, we argue that a
machine role that shapes the assessment belongs in the exam regulations
too, as pre-marking by assistants is admissible only where the procedure
provides for it \citep{bverwg2002_6C7_02}.

\emph{Grading without an examiner.} No study validates it for exams for
credit, and the legal obstacles are two and separate. German examination
law requires the examiner's personal evaluation (L4), which no exception
in data protection law removes. Art. 22 GDPR adds that a solely
automated grade with legal effect needs one of the exceptions of its
paragraph 2 and the safeguards that go with it (L5), which would have to
be shown for the specific use.

\section{Open problems}\label{sec-open}

\emph{Who writes the rubric, and how is a generated one judged?} Expert
rubrics are the bottleneck. Generating them from the reference solution
is the obvious step, but generated rubrics fell short of expert ones in
the comparisons made so far, which generated them without a reference
solution \citep{wang2026generating, zhou2026rubricbench}, the one legal
set generated from expert solutions was only partly reviewed
\citep{oh2026korean}, and whether expert review can close the gap at
acceptable cost is open; rubrics refined against human-scored answers
matched detailed human-written ones in essay scoring
\citep{harada2025automated}, with the caveat of S5. A study of generated
legal rubrics must validate three things apart: the rubric itself (does
it measure the performance the task tests?), its application to real
scripts (does the judge apply it correctly?), and the whole workflow
with the examiner in it (does the final grade improve?). It needs strong
baselines (an independent expert sheet, generated rubrics before and
after expert revision, and a judge without rubric), a split by task
rather than by script, the expert sheet as a reference rather than the
only right answer, so that deviations are classified before they count
as errors, and endpoints beyond correlation: wrong passes and fails,
criterion errors, alternatives and follow-on errors, robustness to the
probes of S3 and the examiner's time. Our companion study on one German
exam is the first step on this programme \citep{nagl2027transformation}.

\emph{Holistic quality.} Some legal quality, such as weighing, judgment
and persuasiveness, may resist criteria fixed in advance
\citep{sadler2009indeterminacy, yang2026judgmentbench}. Hybrids with a
bounded, reasoned overall component or a pairwise component
(Section~\ref{sec-design}) need testing.

\emph{Teaching to the rubric.} Students given rubric feedback may learn
the rubric, not the law, as prompts optimized against a judge learn its
disposition \citep{elganayni2026exploiting}; whether skill or only
scores improve is open.

\section{Limitations}\label{sec-limitations}

Our search is structured but not exhaustive, and several of the most
relevant studies are 2026 preprints. The ratings in Table
\ref{tab:paradigms} are one author's reading of uneven evidence, much of
it from outside law. Several entries in Table \ref{tab:studies} share
data or instruments (marked \emph{reused}) and BenGER is our own, so the
number of papers overstates the number of independent confirmations.
German law is the worked example, and other jurisdictions constrain
grading differently. Five cited works are our own, one of them in
preparation
\citep{nagl2026bengerdata, nagl2026benger, nagl2025courtpressger, elganayni2026exploiting, nagl2027transformation},
and the author works on rubric judges, which may bias the reading
towards them. To limit this, every study, our own included, is coded on
the same fields in Table \ref{tab:studies}.

\section{Conclusion}\label{sec-conclusion}

Evidence from law, education and NLP shows that a rubric written for the
task is the strongest lever for agreement with human graders, and
examination law shows why rubric-centred grading is the family to build
on. The evidence does not yet show that any judge credits defensible
alternatives, counts a follow-on error once, resists manipulation, or is
corrected by examiners when confidently wrong. The trustworthy rubric
judge is our answer: seven safeguards, each tied to a legal requirement,
and a test programme for what is not yet shown. Today it is best placed
for feedback and benchmarks, and as an assistant, not a replacement,
where credit is at stake.

\begin{acks}

This work is generously funded by the Daimler Benz Foundation as part of
project TITAN (\emph{Technologische Intelligenz zur Transformation,
Automatisierung und Nutzerorientierung des Justizsystems}). Claude Opus
5.5, Claude Fable 5.1 and GPT-6 Astra proposed and retrieved candidate
sources, drafted and edited text, and checked every cited claim and
number against its source before submission; the author read the
sources, decided every claim and takes full responsibility for all text.

\end{acks}

\section*{References}

\renewcommand{\bibsection}{}
\bibliography{references.bib}


\end{document}